\documentclass[letterpaper]{article} 
\usepackage{aaai2026}  
\usepackage{times}  
\usepackage{helvet}  
\usepackage{courier}  
\usepackage[hyphens]{url}  
\usepackage{graphicx} 
\usepackage{natbib}  
\usepackage{caption} 
\usepackage{algorithm}
\usepackage{algorithmic}
\usepackage{placeins}

\usepackage{newfloat}
\usepackage{listings}
\DeclareCaptionStyle{ruled}{labelfont=normalfont,labelsep=colon,strut=off} 
\floatstyle{ruled}
\newfloat{listing}{tb}{lst}{}
\floatname{listing}{Listing}
\title{Cheating Stereo Matching in Full-scale: Physical Adversarial Attack against Binocular Depth Estimation in Autonomous Driving}
\author {
	Kangqiao Zhao\textsuperscript{\rm 1}\equalcontrib,
	Shuo Huai\textsuperscript{\rm 1}\equalcontrib,
	Xurui Song\textsuperscript{\rm 2},
	Jun Luo\textsuperscript{\rm 1}\thanks{Corresponding author}
}
\affiliations {
	\textsuperscript{\rm 1}College of Computing and Data Science, Nanyang Technological University, Singapore\\
	\textsuperscript{\rm 2}S-Lab, Nanyang Technological University, Singapore\\
	\{kangqiao.zhao, shuo.huai, junluo\}@ntu.edu.sg,  song0257@e.ntu.edu.sg
}

\usepackage{bibentry}

\usepackage{multirow}
\usepackage{booktabs}
\usepackage{amsmath}
\usepackage{amssymb}

\begin{document}

\maketitle

\begin{abstract}
Though deep neural models adopted to realize the perception of autonomous driving have proven vulnerable to adversarial examples, known attacks often leverage 2D patches and target mostly monocular perception. Therefore, the effectiveness of Physical Adversarial Examples (PAEs) on stereo-based binocular depth estimation remains largely unexplored. To this end, we propose the first texture-enabled physical adversarial attack against stereo matching models in the context of autonomous driving. Our method employs a 3D PAE with global camouflage texture rather than a local 2D patch-based one, ensuring both visual consistency and attack effectiveness across different viewpoints of stereo cameras. To cope with the disparity effect of these cameras, we also propose a new 3D stereo matching rendering module that allows the PAE to be aligned with real-world positions and headings in binocular vision. We further propose a novel merging attack that seamlessly blends the target into the environment through fine-grained PAE optimization. It has significantly enhanced stealth and lethality upon existing hiding attacks that fail to get seamlessly merged into the background. Extensive evaluations show that our PAEs can successfully fool the stereo models into producing erroneous  depth information.
\end{abstract}


\section{Introduction}
\begin{figure*}[t]
	\centering
	\includegraphics[trim=0 10 0 0, clip, width=0.95\textwidth]{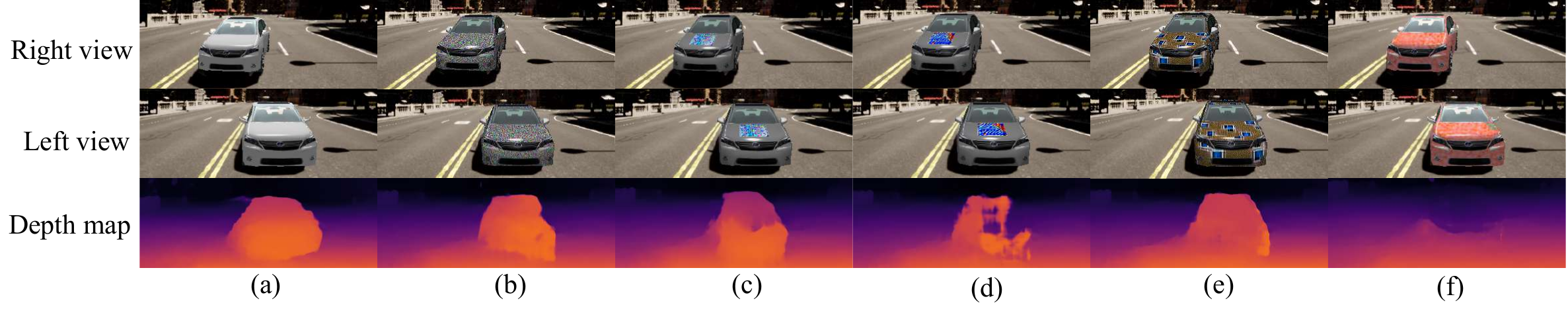}
	\caption{Comparison between existing PAEs and our 3D texture-enabled \textit{merging attack} PAEs in CARLA simulator~\cite{dosovitskiy2017carla}. (a) Benign right and left camera images with the corresponding depth map; (b) Random texture; (c), (d) 2D adversarial patches from~\cite{liu2024physical} and~\cite{cheng2021towards}, respectively, applied to 3D vehicle. (e) MDE Adversarial texture from~\cite{zheng2024physical}; (f) Our 3D \textit{merging attack} adversarial texture.}
	\label{fig:front_view}
\end{figure*}

Recent studies have revealed the vulnerabilities of deep neural perception models for \textit{Autonomous Driving} (AD) against physical adversarial attacks~\cite{
	lovisotto2021slap,zhu2021can,cao2021invisible}. These attacks rely on physically deployable artifacts, also known as \textit{Physical Adversarial Example}s (PAEs)~\cite{eykholt2018robust,thys2019fooling}, to affect the vision features captured by the neural networks and then manipulate their predictions. Currently, most PAEs are in the form of 2D patches that affect only a limited fraction of an object, and they often target only \textit{Monocular Depth Estimation} (MDE) models~\cite{cheng2022physical,guesmi2023aparate,wong2021stereopagnosia,cheng2021towards}. Consequently, these threats to AD might be readily bypassed in real-world 3D scenarios with the support of stereo-based \textit{Binocular Depth Estimation} (BDE)~\cite{xu2020aanet,wang2019pseudo,li2019stereo}.

In fact, attacks on BDE driven by Stereo Matching (SM), a depth estimation method leveraging visual disparity, remain very limited. Sharply different from the PAEs for MDE, attacking SM models presents unique challenges due to its reliance on the physical 3D disparity between two images, rather than only on the 2D image pixels captured by a single camera; it is this uniqueness that invalidates most of MDE attacks under SM-BDE. Although a few proposals~\cite{liu2024physical,cheng2021towards} have explored attacks on stereo matching, they mainly rely on 2D patches  overlaid in \textit{image space}, which not only limits their effectiveness to narrow viewpoints, but also breaks the physical disparity consistency required by real-world systems,  rendering them ineffective under physical deployment.
Meanwhile, SM-BDE is increasingly adopted in AD, with dedicated datasets available to provide stereo data for training SM-BDE models~\cite{geiger2012we, mayer2016large}.
Therefore, it makes sense to consider full-scale 3D PAEs that are effective against SM-BDE, because only such PAEs can substantiate their realistic threat to real-world AD. To generate such PAE, one would need a global 3D camouflage texture to cover the whole surface of a target object,
  because such a texture can significantly increase the affected range within camera views and thus achieve a full-scale adversarial influence on the target.

Adapting PAEs to SM-BDE may also enable more meaningful attacks. Existing attacks on depth estimation primarily focused on hiding objects by pushing their estimated depth to infinity~\cite{guesmi2023aparate,zheng2024physical,cheng2022physical}. As a result, such attacks fail to really hide any objects, as the contour of the object remains, totally exposing the attackers' intention. A more meaningful attack, as one may intuitively expect, should involves adjustable factors for depth fine-tuning, so that attackers can completely adapt the depth of different parts of the target object (including contours) into the surroundings, which we refer to as \textit{merging attack}. Nonetheless, realizing the 3D 
 PAE faces a great challenge under SM-BDE, as each pair of rendered images must precisely align with the physical geometry of stereo cameras in the real world. In addition to making PAEs viable under SM-BDE, a qualified design should also remain robust to varying viewpoints (e.g., position and heading) of the SM camera pair, as well as to diversified external environments (e.g., lighting and weather conditions).

To address the fundamental limitation of image-space attacks under stereo vision, we propose a novel stereo-consistent rendering module that enables physically realizable adversarial camouflage. We leverage 3D object detection to precisely anchor a full-surface adversarial model onto the real-world object with accurate pose alignment. This ensures that the rendered adversarial object produces disparity-consistent projections across both stereo views, a property essential for fooling depth perception in real-world SM-BDE systems.
To realize \textit{merging attack}, we incorporate depth around the contour of the target object to optimize our adversarial camouflage texture. This enables us to understand the contextual information around the object and separate the object into distinct regions with tailored optimization strategies. Compared to uniformly treating the entire object as an optimization target, this approach allows us to make more diversified adjustments.

Additionally, our PAE optimization accounts for diverse environmental variations, including changes in distances, rotations lighting, and weather conditions. Figure~\ref{fig:front_view} presents our \textit{merging attack} PAEs compared to prior works~\cite{liu2024physical, cheng2021towards, zheng2024physical}. The results demonstrate that, under real-world 3D stereo matching constraints, only our method remains effective.
Our main contributions are summarized as follows:

\begin{itemize}
	\item We introduce the first texture-enabled physical adversarial attack against stereo matching models in AD. Our PAE is in form of a 3D camouflage texture attached to the surface of the target object, which can affect object's projection on stereo cameras across various viewpoints.
	\item  We design a novel stereo-aligned 3D rendering module that integrates our adversarial object precisely into the scene using 3D object detection, ensuring faithfully preserving real-world binocular disparity.
	\item We propose a novel and enhanced hiding attack method: \textit{merging attack}, which uses  fine-grained PAE optimization to seamlessly blend the target object into the background, achieving superior stealth and undetectability.
\end{itemize}
We evaluate our attack in both digital simulations and real-world stereo captures, demonstrating strong generalization under varying lighting, weather, and viewpoints. When deployed in a full AD perception stack, our textures cause critical downstream failures, including missed obstacles and faulty path planning, highlighting serious safety risks.

\begin{figure*}[!t]
	\centering
	\includegraphics[width=0.95\linewidth]{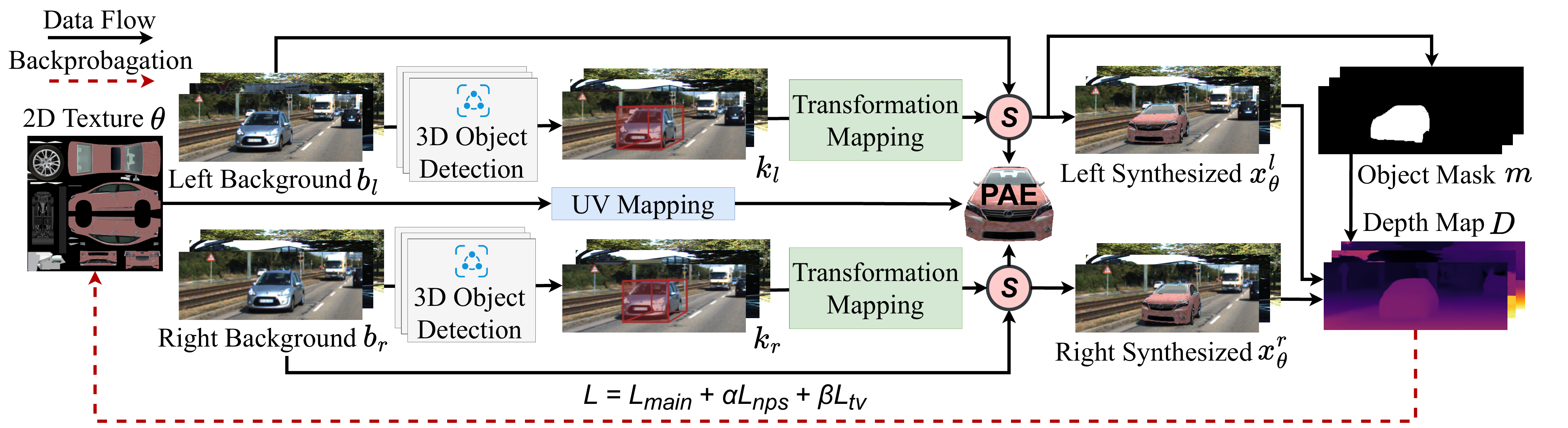}
	\caption{{Overview} of our attack against SM-BDE models. The adversarial camouflage texture $\theta$ is mapped on the PAE and synthesized with the backgrounds ($b_l$ and $b_r$) using a differentiable renderer $R$ with rendering configurations ($k_l$ and $k_r$).  The optimization of $\theta$ is driven by backpropagation, guided by a loss function tailored to the objectives of the \textit{merging attack}.}
	\label{fig:overview}
\end{figure*}

\section{Related Work and Background}

\paragraph{Deep Stereo Matching.}
SM-BDE has received significant attention for its reliability and efficiency in depth estimation for AD~\cite{apollo,waymo,mobileye}. Early works~\cite{zagoruyko2015,zbontar2016} leverage deep networks for separate feature extraction and matching. Subsequent advances include PSMNet~\cite{chang2018}, which incorporates pyramid pooling and 3D CNNs; GA-Net~\cite{zhang2019ga}, which introduces attention-based cost aggregation; and RAFT-Stereo~\cite{lipson2021raft}, which uses recurrent units for multi-level refinement.
Recent works explore cascaded recurrent networks, such as CREStereo~\cite{li2022practical}, and focus on robust generalization and arbitrary-scale disparity, as demonstrated by AnyStereo~\cite{liang2024any}. Given the rapid development of SM-BDE in AD, addressing its security has become critical.

\paragraph{Physical Adversarial Attack.}
Adversarial attacks exploit vulnerabilities in DNN by introducing crafted perturbations to input data, causing the model to make incorrect predictions~\cite{goodfellow2014explaining, madry2017towards}. PAEs have shown real-world effectiveness~\cite{kurakin2018adversarial,eykholt2018robust}, sparking broad interest in safety-critical domains such as AD. They have been studied extensively across key AD tasks, including 3D and 2D object detection~\cite{
	tu2020physically,abdelfattah2021towards,cao2021invisible}, object tracking~\cite{schmalfuss2022perturbation,ranjan2019attacking}, object segmentation~\cite{zhu2021adversarial,xie2017adversarial},
and MDE~\cite{cheng2022physical,
	guesmi2023aparate,wong2021stereopagnosia}.
Recent works~\cite{liu2024physical, cheng2021towards} also explored physical attacks on SM-BDE using 2D PAE but oversimplify stereo image relationships by assuming a direct shift between left and right views, neglecting stereo viewpoint differences. While this assumption may be efficient for digital 2D patch attacks, it fails to generalize to the complexities of real-world 3D scenarios, as shown in Figure~\ref{fig:front_view}. 
However, our 3D camouflage texture as PAE can conform to the parametric constraints of stereo cameras and achieve state-of-the-art attack effectiveness across diverse viewpoints on SM models.

\section{Methodology}

\subsection{Problem Definition}

In this paper, we aim to generate a physically deployable adversarial texture \( \theta \), which can be applied to a real-world object (e.g., a vehicle), such that when observed by a stereo camera, the object can not be perceived by SM-BDE models. This attack, which we refer to as the \textit{merging attack}, misleads the SM-BDE model into assigning background depth to the object, reducing its visibility in the final depth map.

Unlike prior approaches that rely on physically unrealistic perturbations or 2D patches that fail under varying viewpoints, our method optimizes a static adversarial texture that can be physically applied (e.g., printed) to the vehicle surface. Once deployed, this texture remains effective across diverse scenes, viewpoints, and lighting conditions, without any online adjustment or access to stereo inputs.

To simulate real-world stereo perception during training, we use a differentiable rendering pipeline. Given a stereo background image pair \( b = (b^l, b^r) \) and a detected object, we render a textured 3D mesh \( O \) with adversarial texture \( \theta \) under a stereo camera configuration \( k = (k_l, k_r) \), using a renderer \( \mathrm{R} \). The rendered object is then composited with the original background images through a synthesis module \( \mathrm{S} \):
\begin{equation}
	x_\theta = \mathrm{S}(\mathrm{R}(O, \theta, k), b, m),
\end{equation}
where \( m \) is the object mask used for blending. The resulting image pair \( x_\theta = (x_\theta^l, x_\theta^r) \) simulates how the physical object would appear in both stereo views with \(\theta\).
We then optimize \( \theta \) such that the SM-BDE model \( F \) produces depth predictions close to a background-informed target depth \( d_t \):
\begin{equation}
	\textstyle \theta = \arg\min_{\theta} \mathcal{L}(F(x_\theta^l, x_\theta^r), d_t).
\end{equation}
After optimizing, the result is a geometry-aware, physically realizable adversarial texture \( \theta \) that deceives stereo depth models under real-world deployment.

\subsection{Stereo-Aligned 3D Rendering}

While physical adversarial attacks have been widely explored in MDE, their extension to  SM-BDE  is far from straightforward. MDE models infer depth from single image, and adversarial rendering in such settings can be flexibly adjusted, as long as the generated image appears visually plausible. In contrast, SM models depend fundamentally on binocular disparity between two images captured from rigid, spatially coupled viewpoints, which imposes strict physical constraints on how adversarial examples must be rendered.

Specifically, a successful stereo physical attack must ensure that: i) the adversarial object maintains geometrically consistent appearances across both left and right views; ii) the surrounding background context remains coherent in both images to support reliable disparity computation; and iii) the two camera viewpoints follow physically accurate stereo baselines, rather than synthetic approximations. To address these challenges, we introduce a stereo rendering configuration framework, parameterized by $k$, which defines a complete physical stereo setup, inlcuding camera intrinsics, extrinsics, and object placement. This governs how adversarial scenes are rendered. Unlike monocular rendering pipelines, our formulation ensures disparity-consistent dual-view generation, forming the geometric foundation for physically realizable, stereo-consistent adversarial attacks. This design enables us to, for the first time, align adversarial optimization with the physical constraints of real-world stereo-based perception systems in autonomous driving.

To generate physically consistent stereo AEs, as shown in Figure~\ref{fig:overview}, we begin by selecting a stereo image pair from a real-world scene that contains a visible vehicle. Using this pair, we first apply a 3D object detection model to obtain the vehicle’s pose and size, represented by a 3D bounding box:
\begin{equation}
	\mathrm{bbox} = \left\{t_x,t_y,t_z,t_l,t_w,t_h,t_r,t_c\right\},
	\label{eq:3d_detection}
\end{equation}
where \((t_x,t_y,t_z)\) denotes the bounding box’s center position in the world coordinate system, \((t_l,t_w,t_h)\) represent the bounding box dimensions (length, width, and height), \(t_r\) is the heading angle (rotation around the y-axis), and \(t_c\) specifies the object category. As the render camera is set to always look toward the center of the adversarial object, we adopt a spherical coordinate system to parameterize the rendering viewpoint as $k = \{\mathrm{dist}, \mathrm{elev}, \mathrm{azim}\}$, computed as:
\begin{equation}
	\textstyle
	\begin{array}{rl}
		\textstyle \mathrm{dist} &= \sqrt{(c_x - t_x)^2 + (c_y - t_y)^2 + (c_z - t_z)^2},\\
		\textstyle \mathrm{elev} &= \arctan\left(\frac{c_y - t_y}{\sqrt{(c_x - t_x)^2 + (c_z - t_z)^2}}\right),\\
		\textstyle \mathrm{azim} &= \arctan\left(\frac{c_x - t_x}{c_z - t_z}\right).
	\end{array}
\end{equation}
We parameterize the stereo viewpoints as $(k_l, k_r)$ for the left and right cameras, respectively, based on their 3D positions $(c_x, c_y, c_z)$ obtained from stereo calibration. Using these configurations, we independently render a 3D adversarial vehicle with optimized texture $\theta$ into both views, ensuring disparity-consistent appearance. Our method replaces the original object with an adversarial mesh, preserving stereo alignment and depth consistency across the scene.

\subsection{Merging Attack Texture Generation}

In real-world images, objects farther from the camera appear smaller and exhibit lower texture detail. Leveraging this property, we aim to optimize the adversarial texture such that the vehicle visually and geometrically blends into its background. We focus on aligning the vehicle with the surrounding background rather than nearby foreground objects. 
However, this background similarity is not uniform across the adversarial vehicle: the lower regions of the vehicle, which are closer to the ground, tend to align well with the nearby background, while the upper parts show larger depth gaps. This motivates a region-aware optimization strategy that adapts different parts of the object to their respective depth contexts. Specifically, to ensure effective \textit{merging attack}, we design a three-step process: boundary depth extraction, region segmentation, and texture optimization.

\begin{figure}[!t]
	     \centering
	     \includegraphics[width=0.95\linewidth]{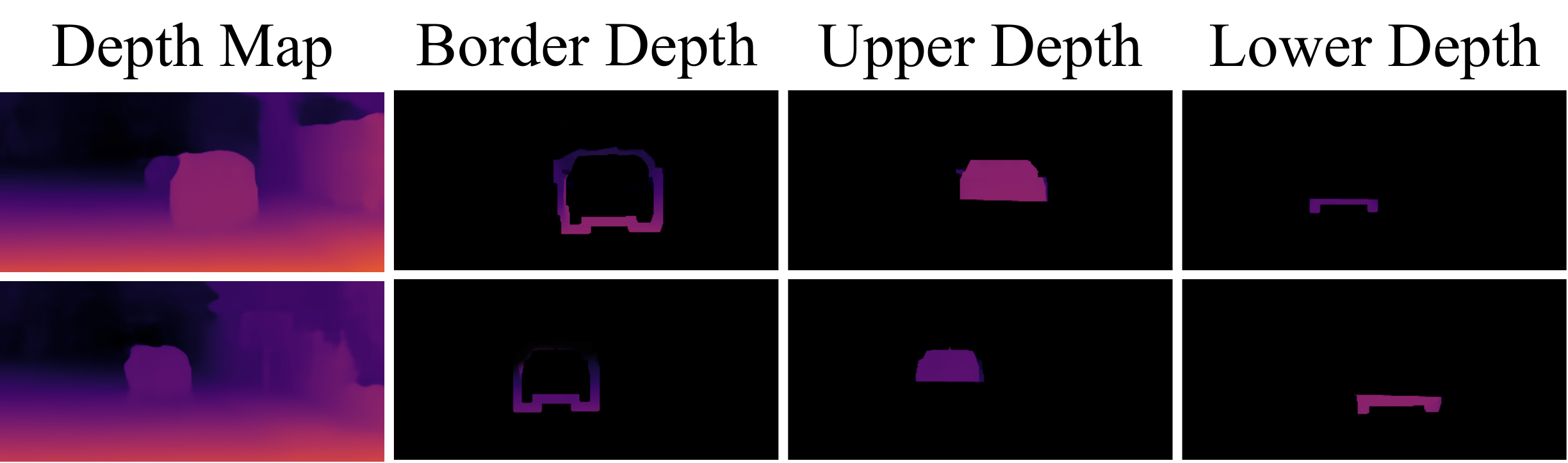}
	     \caption{Depth results of boundary and each segmentation.}
	     \label{fig:mask_seg_main}
	 \end{figure}

During boundary depth extraction, we obtain the surrounding background depth from the scene's depth map $d$ by expanding the object mask $m$ (see Figure~\ref{fig:mask_seg_main}). Applying max pooling with kernel size $p_k \times p_k$ enlarges $m$ to cover nearby regions, and subtracting the original mask yields the boundary mask $m_{bg}$. We then extract the background depth $d_{bg}$ via element-wise multiplication with $d$:
\begin{equation}
	m_{bg} = \mathrm{Maxpool}(m) - m, \ \ \ d_{bg} = d \cdot m_{bg}.
\end{equation}

Next, in the region segmentation stage, we divide  the adversarial vehicle’s depth map horizontally to optimize different regions for varying surrounding depths. To balance simplicity and effectiveness, we divide the map into upper and lower regions,
 as shown in Figure~\ref{fig:mask_seg_main}. 
To anchor this division, we compute the mean background depth from the previously extracted boundary map, $d_{bg}^{\textit{avg}} = \sum d_{bg} / \sum m_{bg}$, to split the vehicle depth. We then locate two reference points on the left and right sides of the object boundary whose depths are closest to $d_{bg}^{\textit{avg}}$. These points define a horizontal contour across the object surface that approximately aligns with the background depth plane, guiding the split between regions. This segmentation enables the upper and lower parts of the texture to be optimized toward distinct depth priors, enhancing the effectiveness of\textit{ merging attack}. 

Finally, during texture optimization,  we introduce a \textit{region-aware depth alignment loss} that guides the adversarial texture to match the depth statistics of the surrounding background, aiming to achieve seamless blending between the adversarial object and its environment. Specifically, we compute the average depth of the object and the surrounding region in both the upper and lower segments defined by the horizontal partition. The main \textit{merging loss} is formulated as:
\begin{equation} 
	\mathcal{L}_{\text{merge}}(\theta) = \mathrm{MSE}(d^{up}_{\textit{obj}}, d^{up}_{\textit{bg}}) + \mathrm{MSE}(d^{bt}_{\textit{obj}}, d^{bt}_{\textit{bg}}),
	\label{eq:merge_loss}
\end{equation}
where \( d^{up}_{\textit{obj}} \) and \( d^{bt}_{\textit{obj}} \) denote the average depth of the object's upper and bottom regions, respectively, and \( d^{up}_{\textit{bg}} \), \( d^{bt}_{\textit{bg}} \) are the corresponding local background depths extracted around the object boundary. By independently optimizing each region to align with its local background depth,
the attack blend the adversarial vehicle  into the scene, resulting in a more concealed and less detectable adversarial effect.

To ensure that the adversarial texture is both physically printable and visually smooth, we further incorporate two regularization terms: a total variation (TV) loss ($\mathcal{L}_{\text{tv}}$)~\cite{mahendran2015understanding} that suppresses high-frequency noise, and a non-printability score (NPS) loss ($\mathcal{L}_{\text{nps}}$)~\cite{sharif2016accessorize} that encourages the use of printer-reproducible colors. The final loss function is:
\begin{equation} 
	\mathcal{L}(\theta) = \mathcal{L}_{\text{merge}}(\theta) + \alpha \mathcal{L}_{\text{nps}}(\theta) + \beta \mathcal{L}_{\text{tv}}(\theta),
	\label{eq:total_loss}
\end{equation}
where \(\alpha\) and \(\beta\) are hyperparameters balancing the merging objective, printability, and smoothness.

\paragraph{Appearing Attack.} Our \textit{merging attack} represents an advanced form of hiding attack, designed to reduce the visibility of the target vehicle. Complementary to \textit{merging attack}, we also introduce a strategy, \textit{appearing attack}, which aims to make the target vehicle highly conspicuous. This could forces surrounding vehicles to engage in sudden braking maneuvers, potentially causing accidents and unexpected disruptions in traffic flow. To achieve this objective, we define the loss function $\mathcal{L}_{\text{appear}}$ for texture optimization, which seeks to minimize the perceived depth within the target mask  $m$, thereby minimizing the visual distance between the adversarial object and the camera. This is achieved by minimizing the MSE between $d_{obj}$ and $D_{max}$, where $D_{max}$ is set according to model's configuration for upper limit:
\begin{equation} 
	\mathcal{L}_{\text{appear}}(\theta) = \mathrm{MSE}(d_{obj},D_{max}). 
\end{equation}

The total loss $\mathcal{L}(\theta)$ for our \textit{appearing attack} follows a formulation similar to that of the \textit{merging attack}, with $\mathcal{L}_{\text{merge}}(\theta)$ in Eq.~\eqref{eq:total_loss} replaced by $\mathcal{L}_{\text{appear}}(\theta)$.

\begin{table*}[!t]
	\centering
	\small
	\setlength{\tabcolsep}{3.2pt}
	\renewcommand{\arraystretch}{1.1}
	\begin{tabular}{lccc ccc ccc ccc ccc}
		\toprule
		\multirow{2}{*}{\textbf{Method}} 
		& \multicolumn{3}{c}{\textbf{PSMNet}} 
		& \multicolumn{3}{c}{\textbf{GA-Net}} 
		& \multicolumn{3}{c}{{\textbf{RAFT-Stereo}}}
		& \multicolumn{3}{c}{{\textbf{CREStereo}}}
		& \multicolumn{3}{c}{{\textbf{AnyStereo}}} \\
		\cmidrule(r){2-4} \cmidrule(r){5-7} \cmidrule(r){8-10} \cmidrule(r){11-13} \cmidrule(r){14-16}
		& $\mathcal{E}_\text{blend}\downarrow$ & $\mathcal{E}_\text{cover}\uparrow$ & $\mathcal{E}_\text{sys}\uparrow$
		& $\mathcal{E}_\text{blend}\downarrow$ & $\mathcal{E}_\text{cover}\uparrow$ & $\mathcal{E}_\text{sys}\uparrow$
		& $\mathcal{E}_\text{blend}\downarrow$ & $\mathcal{E}_\text{cover}\uparrow$ & $\mathcal{E}_\text{sys}\uparrow$
		& $\mathcal{E}_\text{blend}\downarrow$ & $\mathcal{E}_\text{cover}\uparrow$ & $\mathcal{E}_\text{sys}\uparrow$
		& $\mathcal{E}_\text{blend}\downarrow$ & $\mathcal{E}_\text{cover}\uparrow$ & $\mathcal{E}_\text{sys}\uparrow$ \\
		\midrule
		Benign & 0.631 & 0.013 & 0 & 0.641 & 0.012 & 0 & 0.786 & 0.012 & 0 & 0.677 & 0.017 & 0 & 0.572 & 0.093 & 0 \\
		Random & 0.677 & 0.052 & 0 & 0.680 & 0.031 & 0 & 0.793 & 0.024 & 0 & 0.684 & 0.029 & 0 & 0.580 & 0.022 & 0 \\
		PASM & 0.475 & 0.154 & 0.13 & 0.411 & 0.088 & 0.12 & 0.502 & 0.148 & 0.07 & 0.431 & 0.094 & 0.15 & 0.471 & 0.124 & 0.15 \\
		Adv-DM & 0.510 & 0.176 & 0.04 & 0.449 & 0.075 & 0.12 & 0.614 & 0.143 & 0.05 & 0.444 & 0.077 & 0.17 & 0.480 & 0.119 & 0.09 \\
		\midrule
		\textbf{Ours} & \textbf{0.058}  & \textbf{0.553} & \textbf{0.74}
		& \textbf{0.069}  & \textbf{0.588} & \textbf{0.69}
		& \textbf{0.082}  & \textbf{0.571} & \textbf{0.62}
		& \textbf{0.071}  & \textbf{0.598} & \textbf{0.70}
		& \textbf{0.056}  & \textbf{0.576} & \textbf{0.76} \\
		\bottomrule
	\end{tabular}
	\caption{Comparison results between existing attacks and our \textit{merging attack} against various target models.}
	\label{tab:overall_comparison}
\end{table*}

\section{Experiments}

\subsection{Experiment Setup}

\paragraph{Implementation Details.}
We train our adversarial textures using the KITTI2015 dataset~\cite{geiger2012we}, a widely adopted benchmark for AD that provides   stereo image pairs. The training samples cover diverse vehicle poses, scales, and backgrounds, promoting generalization across real-world scenes.
To enhance the robustness of adversarial textures under diverse environmental conditions, we adopt Expectation over Transformation (EoT)~\cite{athalye2018synthesizing} during training. Specifically, we randomly perturb the position of a point light source within the range \([-3, 3]\) meters along each axis, and uniformly sample ambient light intensity from \([0.3, 0.9]\). To simulate challenging weather such as rain and fog, we additionally inject random Gaussian noise into the rendered images during training.
For boundary depth extraction, the kernel size $p_k$ is set to $40$.
We optimize the adversarial texture using a differentiable renderer and train for 100 epochs with the Adam optimizer~\cite{kingma2014adam},  with an initial learning rate of 0.01, scheduled via cosine decay to a minimum of $1e-4$. The loss coefficients are set to $\alpha=5$ and $\beta=0.1$.
All experiments are performed on a single NVIDIA RTX A5000 GPU.

\paragraph{Evaluation Metrics.} We evaluate our 3D adversarial camouflage attack from both perception and system perspectives using three metrics:
i) Hiding Error ($\mathcal{E}_\text{blend}$), which evaluates how well the adversarial object blends into its surroundings by measuring the proportion of object pixels whose depth significantly deviates from that of the background contour:
\begin{equation}
	\mathcal{E}_\text{blend} = \textstyle\frac{1}{\sum_{m}}\cdot \sum_{(i, j)\in m}\mathbb{I}\left( |d(i,j) - \bar{d}_{\text{bg}}(i)| > \tau \right),
\end{equation}
where $d(i,j)$ is the predicted depth at pixel $(i,j)$, and \( \mathbb{I}(\cdot) \) is the indicator function.  
The contour average depth $\bar{d}_{\text{bg}}(i)$ is computed per row $i$ using pixels selected by     $m_{\text{bg}}$.  $\tau$ is a predefined threshold used to determine various depth differences; we set $\tau = 20 $ in our experiments.
ii) Perturbation Coverage Ratio ($\mathcal{E}_\text{cover}$), which quantifies the fraction of the object region where depth has been altered by the :
\begin{equation}
	\textstyle   \mathcal{E}_\text{cover} = \frac{1}{\sum_m} \cdot \sum_{(i, j)\in m}  \mathbb{I}\left( d^{\text{adv}}(i,j)  \neq d^{\text{benign}}(i,j) \right),
\end{equation}
where  \(d^{\text{adv}}\) and  \( d^{\text{benign}} \) are the depth maps under adversarial and benign   textures, respectively. iii) System-Level Collision Rate ($\mathcal{E}_\text{sys}$), which evaluates downstream safety. We use a full Apollo   perception and planning stack. We define the system-level metric as the proportion of evaluation episodes where the ego vehicle fails to avoid the adversarial object:
\begin{equation}
	\textstyle   \mathcal{E}_\text{sys} = \frac{\text{\# of collisions}}{\text{\# of total runs}}.
\end{equation}

Higher values of \( \mathcal{E}_\text{cover} \)  and \( \mathcal{E}_\text{sys} \) indicate stronger adversarial impact, while lower \( \mathcal{E}_\text{blend} \) reflects better concealment and a more effective merging attack.

For the \textit{appearing attack} (Results in Table~\ref{tab:appear_eval}), we replace $\mathcal{E}_\text{blend}$ with the Mean Depth Shift ($\mathcal{E}_\text{shift}$), which measures how much the adversarial object appears closer to the camera compared to its benign counterpart. A higher value indicates a stronger  effect caused by the adversarial texture:
\begin{equation}
	\textstyle    \mathcal{E}_\text{shift} = \frac{1}{\sum_m} \cdot    \sum_{(i, j)\in m}  \left( d^{\text{adv}}(i,j)  - d^{\text{benign}}(i,j) \right).
\end{equation}

\begin{table}[]
	\centering
	\small
	\setlength{\tabcolsep}{3pt}
	\renewcommand{\arraystretch}{1.1}
	\begin{tabular}{c c cc cc cc}
		\toprule
		\multirow{2}{*}{\textbf{Metric}} & \multirow{2}{*}{\textbf{Tex.}} 
		& \multicolumn{2}{c}{\textbf{Time of Day}} 
		& \multicolumn{2}{c}{\textbf{View Angle}} 
		& \multicolumn{2}{c}{\textbf{Distance (m)}} \\
		\cmidrule(lr){3-4} \cmidrule(lr){5-6} \cmidrule(lr){7-8}
		&  & Midday & Sunset & Side & Front & 12 & 24 \\
		\midrule
		\multirow{2}{*}{$\mathcal{E}_\text{blend} \downarrow$} 
		& Benign  & 0.481 & 0.536 & 0.557 & 0.513 & 0.517 & 0.402\\
		& \textbf{Adv.} & \textbf{0.087} & \textbf{0.067} & \textbf{0.071} & \textbf{0.085} & \textbf{0.074} & \textbf{0.095} \\
		\midrule
		\multirow{2}{*}{$\mathcal{E}_\text{cover} \uparrow$} 
		& Benign  & 0.036 & 0.042 & 0.030 & 0.038 & 0.035 & 0.032 \\
		& \textbf{Adv.} & \textbf{0.519} & \textbf{0.577} & \textbf{0.581} & \textbf{0.520} & \textbf{0.504} & \textbf{0.467} \\
		\bottomrule
	\end{tabular}
	\caption{Real-world evaluation results under varying time of day, viewing angle, and distances.}
	\label{tab:realworld_comparison}
\end{table}

\begin{figure}[!t]
	\centering
	\includegraphics[width=0.95\linewidth]{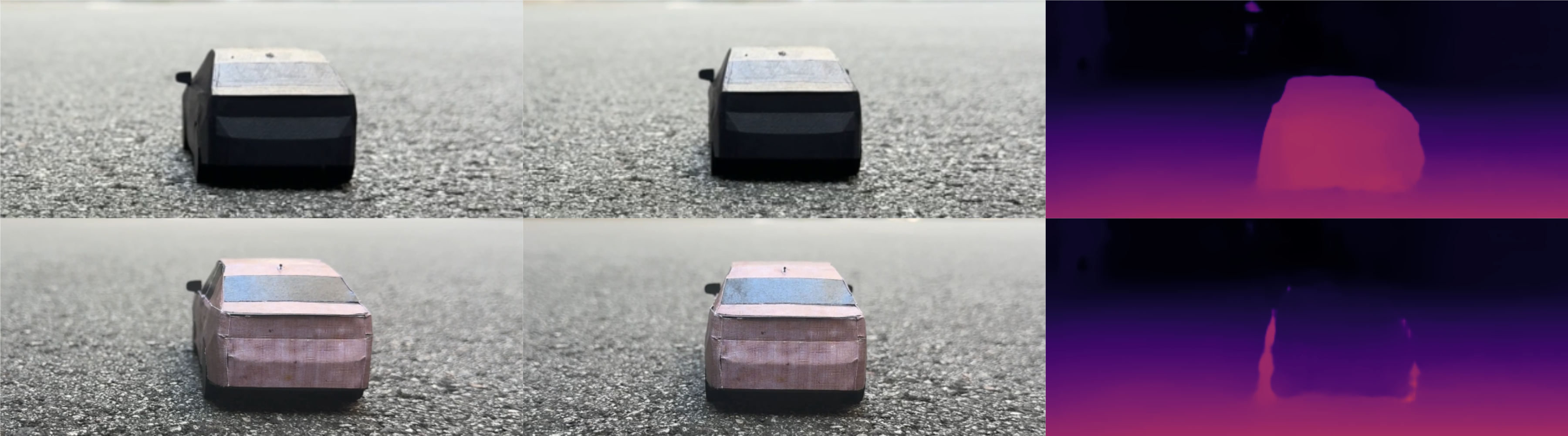}
	\caption{{Visualization of our 3D PAEs in the real world. Top: Benign. Bottom: Adversarial.}}
	\label{fig:phy_attack}
\end{figure}

\begin{figure*}[t]
	\centering
	\includegraphics[width=0.95\linewidth]{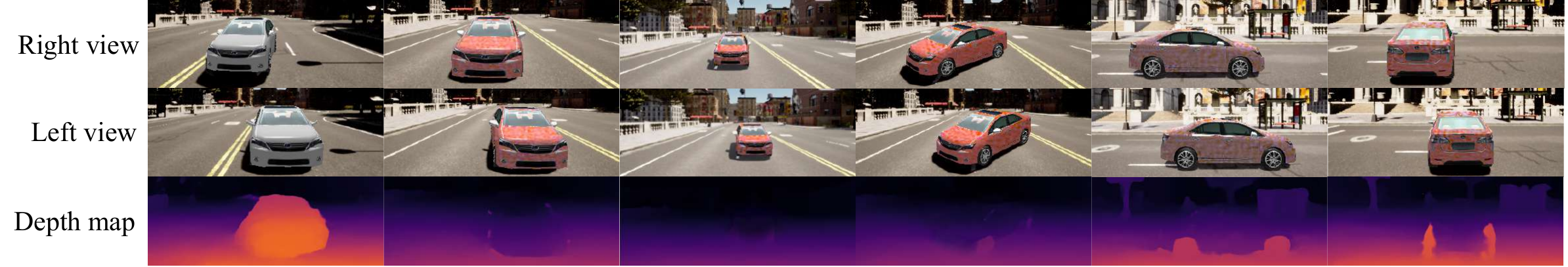}
	\caption{Our 3D texture-PAEs against SM-BDE under different viewpoint variations.}
	\label{fig:distance_angle}
\end{figure*}

\begin{table*}[!t]
	\centering
	\small
	\setlength{\tabcolsep}{4pt}
	\renewcommand{\arraystretch}{1.1}
	\begin{tabular}{lcccc|ccccccccccccc}
		\toprule
		\multirow{2}{*}{\textbf{Method}} & \multirow{2}{*}{\textbf{Metric}} 
		& \multicolumn{3}{c|}{\textbf{Distance (m)}} 
		& \multicolumn{12}{c}{\textbf{Heading Angle ($^\circ$)}} \\
		\cmidrule(lr){3-5} \cmidrule(lr){6-17}
		& & 3–9 & 9–15 & 15–20 
		& 0 & 30 & 60 & 90 & 120 & 150 & 180 & 210 & 240 & 270 & 300 & 330 \\
		
		\midrule
		Benign   & \multirow{5}{*}{$\mathcal{E}_\text{blend} \downarrow$} 
		& 0.684 & 0.628 & 0.481
		& 0.657 & 0.644 & 0.639 & 0.609 & 0.612 & 0.633 & 0.640 & 0.625 & 0.629 & 0.618 & 0.633 & 0.653 \\
		Random   &       
		& 0.704 & 0.681 & 0.489 
		& 0.684 & 0.667 & 0.654 & 0.656 & 0.641 & 0.648 & 0.659 & 0.654 & 0.651 & 0.640 & 0.649 & 0.659 \\
		PASM     &       
		& 0.475 & 0.455 & 0.417 
		& 0.461 & 0.480 & 0.481 & 0.515 & 0.484 & 0.488 & 0.473 & 0.491 & 0.504 & 0.467 & 0.472 & 0.483 \\
		Adv-DM   &       
		& 0.510 & 0.461 & 0.423 
		& 0.516 & 0.534 & 0.541 & 0.578 & 0.536 & 0.524 & 0.520 & 0.542 & 0.554 & 0.585 & 0.547 & 0.536 \\
		\textbf{Stereo-Align}     &       
		& \textbf{0.058} & \textbf{0.069} & \textbf{0.084} 
		& \textbf{0.058} & \textbf{0.062} & \textbf{0.060} & \textbf{0.051} & \textbf{0.059} & \textbf{0.065} & \textbf{0.065} & \textbf{0.063} & \textbf{0.059} & \textbf{0.050} & \textbf{0.057} & \textbf{0.063} \\
		\midrule
		Benign   &  \multirow{5}{*}{$\mathcal{E}_\text{cover} \uparrow$}
		& 0.037 & 0.029 & 0.024 
		& 0.028 & 0.025 & 0.029 & 0.015 & 0.017 & 0.027 & 0.030 & 0.025 & 0.020 & 0.017 & 0.021 & 0.026 \\
		Random   &       
		& 0.053 & 0.041 & 0.034 
		& 0.048 & 0.045 & 0.039 & 0.033 & 0.036 & 0.046 & 0.048 & 0.041 & 0.037 & 0.032 & 0.037 & 0.044 \\
		PASM     &       
		& 0.183 & 0.165 & 0.093 
		& 0.132 & 0.120 & 0.111 & 0.094 & 0.107 & 0.114 & 0.144 & 0.121 & 0.105 & 0.099 & 0.112 & 0.126 \\
		Adv-DM   &       
		& 0.211 & 0.176 & 0.098 
		& 0.146 & 0.128 & 0.122 & 0.107 & 0.129 & 0.129 & 0.132 & 0.125 & 0.118 & 0.102 & 0.117 & 0.125 \\
		\textbf{Stereo-Align}     &       
		& \textbf{0.585} & \textbf{0.532} & \textbf{0.468} 
		& \textbf{0.496} & \textbf{0.533} & \textbf{0.548} & \textbf{0.595} & \textbf{0.543} & \textbf{0.536} & \textbf{0.501} & \textbf{0.523} & \textbf{0.544} & \textbf{0.562} & \textbf{0.547} & \textbf{0.528} \\
		\bottomrule
	\end{tabular}
	\caption{
		Evaluation of attack robustness across various distances and a full range of heading angles.  Our method consistently achieves higher $\mathcal{E}_\text{cover}$ and lower $\mathcal{E}_\text{blend}$ under varied viewpoints.
	}
	\label{tab:distance_angle}
\end{table*}

\paragraph{Compared Methods \& Target SM Models.} 
We compare our method with the two most relevant physical attacks designed against SM-BDE: PASM~\cite{liu2024physical} and {Adv-DM}~\cite{cheng2021towards}. In addition, we include two baselines textures: \textit{benign} (neutral textures) and \textit{random} (random noise). We evaluate all textures on five representative SM-BDE models: PSMNet~\cite{chang2018}, GA-Net~\cite{zhang2019ga}, RAFT-Stereo~\cite{lipson2021raft}, CREStereo~\cite{li2022practical}, and AnyStereo~\cite{liang2024any}, covering diverse cost
volume constructions and model designs.

\subsection{Attack Effectiveness}

\paragraph{Digital Evaluation.} We evaluate the trained textures in the CARLA simulator~\cite{dosovitskiy2017carla} using a world-aligned stereo setup. Adversarial vehicles are placed at distances ranging from 3 to 30 meters and orientations between \(0^\circ\) and \(360^\circ\), under varying lighting and weather conditions in the “Town 10” map.
Table~\ref{tab:overall_comparison} summarizes the evaluation results of our \textit{merging attack} in comparison with state-of-the-art physical attacks across five representative SM-BDE models. All attacks are tested under identical camera baselines and environmental conditions in the CARLA simulator. Our adversarial texture consistently achieves a higher $\mathcal{E}_\text{cover}$, indicating a broader impact on predicted depth. This can be attributed to the larger effective surface coverage and better generalization across object orientations. Moreover, our approach attains the lowest $\mathcal{E}_\text{blend}$  among all methods, suggesting superior blending into the surrounding environment and reduced detectability in depth predictions, both key objectives of hiding attacks. Importantly, when integrated into a full commercial AD pipeline -- Apollo, our attack leads to the highest $ \mathcal{E}_\text{sys}$, indicating a significant increase in missed detection and path planning failures. This highlights the safety risk posed by our physically realizable texture.

\paragraph{Physical Evaluation.}
To assess real-world deployability, we 3D-print a 1:30 scale sedan and physically apply the optimized texture to its surface. A total of 300 stereo image pairs were collected across diverse conditions (e.g., viewing angles, distances, and environments) using two iPhone 14 Pro Max cameras, emulating the KITTI stereo setup. As shown in Table~\ref{tab:realworld_comparison}, our method maintains strong performance across all evaluation metrics under transformation and lighting variations. Notably, $\mathcal{E}_\text{blend}$ remains low, indicating that the adversarial vehicle seamlessly merges into the surrounding under physical conditions. As illustrated in Figure~\ref{fig:phy_attack}, while the benign-textured vehicle maintains clear geometric boundaries in depth maps, our adversarial texture causes the vehicle to disappear into the background.  These results highlight the physical realizability and
robustness of our method.

\subsection{Attack Analysis}

\paragraph{Attack Robustness.}
To evaluate the robustness of our attack under diverse real-world conditions, we conduct extensive tests across varying camera-object distances, viewing angles, and environmental factors such as weather and lighting. As shown in Figure~\ref{fig:distance_angle}, our adversarial textures consistently outperform prior methods across all settings, achieving higher affected region ratios $\mathcal{E}_\text{cover}$ and lower hiding errors $\mathcal{E}_\text{blend}$. These results are further validated by the quantitative metrics summarized in Table~\ref{tab:distance_angle}. While performance generally degrades at longer distances due to reduced visual resolution, our method remains notably more robust to angular variation. Unlike patch-based attacks that depend on frontal visibility, our full-surface adversarial texture maintains effectiveness from oblique viewpoints due to its geometry-aware design. In addition, our attack remains stable under challenging weather conditions such as rain and fog. Even in the worst case, it achieves a hiding error $\mathcal{E}_\text{blend}$ of $0.091$ and an affected region ratio $\mathcal{E}_\text{cover}$ of $0.463$, confirming its resilience in low-visibility scenarios. More settings and results are provided in the \textit{supplementary material}.

\paragraph{Ablation Study.}
We conduct an ablation study to evaluate the effectiveness of the Stereo-Aligned 3D Rendering (SAR) module and the \textit{merging attack} mechanism. We compare adversarial textures generated under four configurations: no module (\textit{None}), \textit{SAR} only, \textit{merging} only, and the full method with both components.
In the \textit{merging only} setting, we define the main loss $\mathcal{L}_{\text{main}}$ as $\mathcal{L}_{\text{merge}}$ and reuse the same rendering parameters $k$ from the left view for the right. For \textit{None} and \textit{SAR only}, we replace $\mathcal{L}_{\text{merge}}$ with a standard hiding loss~\cite{zheng2024physical}.
As shown in Table~\ref{tab:ablation}, removing the physical constraint $\mathcal{L}_{\text{nps}}$ slightly improves performance due to the relaxation of printability limits. However, without enforcing stereo consistency, the texture has little effect on the model’s predictions. In contrast, applying SAR significantly increases the affected region on the depth map (high $\mathcal{E}_\text{cover}$), while omitting the \textit{merging} loss results in a clearly visible vehicle contour (low $\mathcal{E}_\text{blend}$).

\begin{table}[t]
	\centering
	\resizebox{\columnwidth}{!}{%
		\setlength{\tabcolsep}{5pt} 
		\renewcommand{\arraystretch}{1.1}
		\begin{tabular}{cccc cccc}
			\toprule
			\multirow{2}{*}{\textbf{Metric}} & \multicolumn{3}{c}{\textbf{Losses}} & \multicolumn{4}{c}{\textbf{Modules}} \\ \cmidrule(lr){2-4}  \cmidrule(lr){5-8}
			& $\mathcal{L}_{\text{main}}$ & $\mathcal{L}_{\text{nps}}$ & $\mathcal{L}_{\text{tv}}$ & \textit{None} & \textit{SAR} & \textit{Merge} & \textit{Full}  \\ \midrule
			$\mathcal{E}_\text{blend} \downarrow$ & \multirow{2}{*}{\checkmark} & \multirow{2}{*}{}  & \multirow{2}{*}{}  & 0.611 & 0.393 & 0.615 & 0.049  \\
			$\mathcal{E}_\text{cover} \uparrow$ &  &  &  & 0.017 & 0.556 & 0.029 & 0.607 \\  \midrule
			$\mathcal{E}_\text{blend} \downarrow$ & \multirow{2}{*}{\checkmark} & \multirow{2}{*}{\checkmark} & \multirow{2}{*}{}  & 0.631 & 0.403 & 0.611 & 0.051  \\
			$\mathcal{E}_\text{cover} \uparrow$ &  &  &  & 0.015 & 0.541 & 0.024 & 0.587 \\  \midrule
			$\mathcal{E}_\text{blend} \downarrow$ & \multirow{2}{*}{\checkmark} & \multirow{2}{*}{}  & \multirow{2}{*}{\checkmark} & 0.602 & 0.379 & 0.598 & \textbf{0.042}  \\
			$\mathcal{E}_\text{cover} \uparrow$ &  &  &  & 0.020 & 0.582 & 0.031 & \textbf{0.616} \\  \midrule
			$\mathcal{E}_\text{blend} \downarrow$ & \multirow{2}{*}{\checkmark} & \multirow{2}{*}{\checkmark} & \multirow{2}{*}{\checkmark} & 0.685 & 0.411 & 0.665 & 0.058  \\
			$\mathcal{E}_\text{cover} \uparrow$ &  &  &  & 0.015 & 0.515 & 0.019 & 0.573 \\ \bottomrule
		\end{tabular}%
	}
	\caption{Ablation results with different modules and losses.}
	\label{tab:ablation}
\end{table}

\begin{table}[!t]
	\centering
	\resizebox{\columnwidth}{!}{%
		\setlength{\tabcolsep}{5pt}
		\renewcommand{\arraystretch}{1.1}
		\begin{tabular}{cccccc}
			\toprule
			\textbf{Metric} & \textbf{PSMNet} & \textbf{GA-Net} & {\textbf{RAFT}} & {\textbf{CRE}} & {\textbf{AnyStereo}} \\ \midrule
			$\mathcal{E}_\text{cover} \uparrow$   & 0.427   & 0.405  & 0.455 & 0.422  & 0.415       \\ 
			$\mathcal{E}_\text{shift} \uparrow$  & 16.48   & 15.87  & 15.64 & 15.57  & 15.10       \\ \bottomrule
		\end{tabular}%
	}
	\caption{\textit{Appearing attack} against different SM models.}
	\label{tab:appear_eval}
\end{table}

\begin{figure}[!t]
	\centering
	\includegraphics[width=0.95\linewidth]{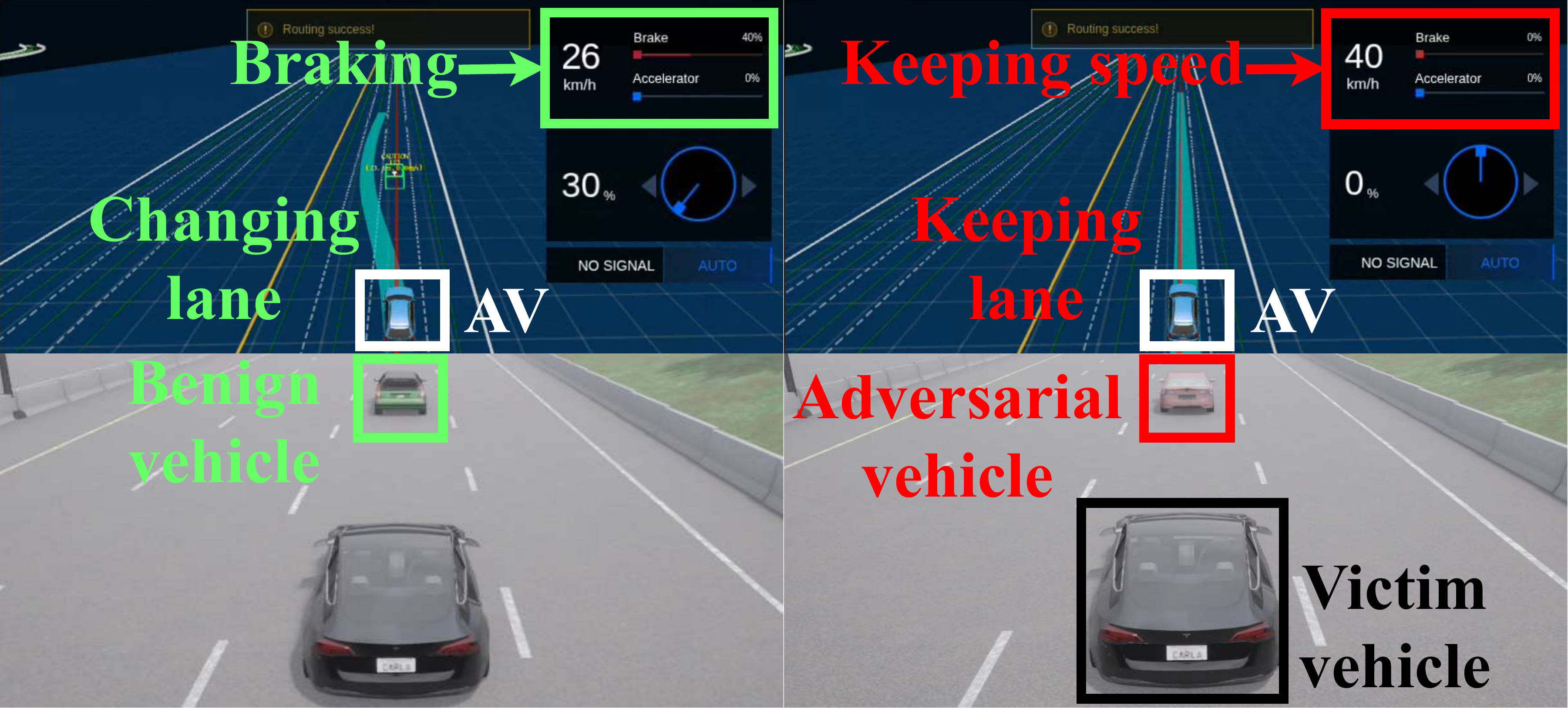}
	\caption{System-level effect of our attack.}
	\label{fig:sys_attack}
\end{figure}

\paragraph{System Evaluation.}
To assess the real-world impact of our attack, we conduct system-level experiments on the open-source autonomous driving platform Baidu Apollo. In each scenario, the adversarial vehicle is placed ahead of the victim AV to simulate realistic traffic. We select representative scenes and feed perception outputs, after decision-level fusion~\cite{zhu2024malicious}, into Apollo's planning module under both benign and adversarial conditions.
In the benign case, the AV detects the vehicle ahead, slows down, and initiates a lane change to avoid collision. In contrast, under attack, the adversarial texture disrupts scene understanding, causing the AV to maintain its speed and lane, ultimately resulting in a rear-end collision. As shown in Figure~\ref{fig:sys_attack}, the AV safely changes lane at a 10\,m distance in the benign setting. However, in the adversarial case, it fails to respond and collides with the target in both tests. These results highlight the severe safety risks posed by stereo-consistent physical attacks.

\begin{figure}[!t]
	\centering
	\includegraphics[width=0.95\linewidth]{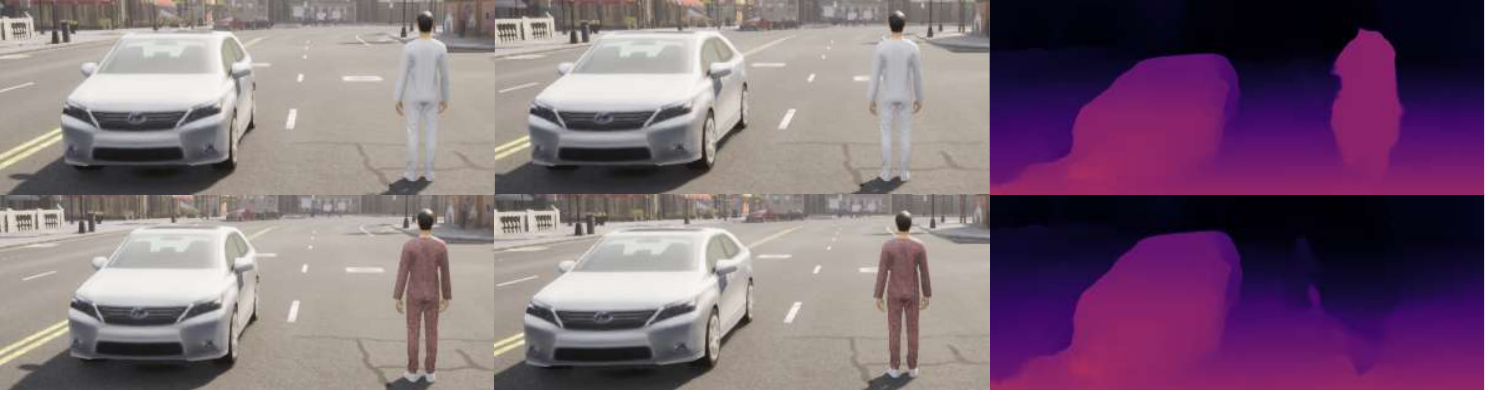}
	\caption{Attack transferability on human model. Top: Benign. Bottom: Adversarial}
	\label{fig:trans_attack}
\end{figure}

\paragraph{Attack Transferability.}
{
	We first evaluate cross-model
	transferability by testing whether adversarial textures trained
	on one stereo model can effectively attack others.
	Textures optimized on PSMNet transfer well to GANet ($\mathcal{E}_\text{blend} = {0.087}$; $\mathcal{E}_\text{cover} = {0.524}$), likely due to similar structures and disparity mechanisms. In contrast, transferability to models with iterative refinement paradigm like RAFT-Stereo ($\mathcal{E}_\text{blend} = {0.128}$; $\mathcal{E}_\text{cover} = {0.451}$) 
	 is relatively weaker}. We further examine cross-category transferability by training adversarial textures on a human mesh. As shown in Figure~\ref{fig:trans_attack}, when an adversarial-textured human model is placed at the same depth as a benign vehicle in the simulation environment, the resulting depth map shows that the adversarial-textured human blends seamlessly into the background.

\begin{figure}[!t]
	\centering
	\includegraphics[width=0.95\linewidth]{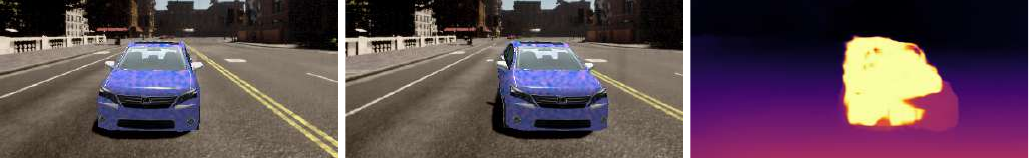}
	\caption{Depth map of appearing attack.}
	\label{fig:app_attack}
\end{figure}

\section{Conclusion}

In this paper, we introduce the first texture-enabled physical adversarial attack against stereo matching-based binocular depth estimation in the context of autonomous driving. To precisely align with the disparity features utilized in stereo matching models, we develop a stereo-synchronized 3D rendering module that ensures view disparity remain consistent with the stereo cameras. This synchronization enables the generation of rendered objects that faithfully replicate real-world 3D transformations. We further propose a novel hiding strategy, termed {merging attack}, which can seamlessly blend the adversarial object into the background, enhancing stealth and reducing detectability. Extensive experiments at both the perception and system levels demonstrate that our attack generalizes across stereo matching models and environmental conditions, effectively compromising the full 3D structure of the target object.

\section*{Acknowledgments}
This research/project is supported by the National Research Foundation, Singapore, under its AI Singapore Programme (AISG Award No: AISG4-GC-2023-006-1B).

\bibliography{aaai2026}

@article{kingma2014adam,
	title={{Adam: A Method for Stochastic Optimization}},
	author={Kingma, Diederik P},
	journal={arXiv preprint arXiv:1412.6980},
	year={2014}
}

@inproceedings{sharif2016accessorize,
	title={{Accessorize to a Crime: Real and Stealthy Attacks on State-of-the-Art Face Recognition}},
	author={Sharif, Mahmood and Bhagavatula, Sruti and Bauer, Lujo and Reiter, Michael K},
	booktitle={Proc. of the 23rd ACM CCS},
	pages={1528--1540},
	year={2016}
}

@inproceedings{mahendran2015understanding,
	title={{Understanding Deep Image Representations by Inverting Them}},
	author={Mahendran, Aravindh and Vedaldi, Andrea},
	booktitle={Proc. of the 28th IEEE/CVF CVPR},
	pages={5188--5196},
	year={2015}
}

@article{goodfellow2014explaining,
  title={{Explaining and Harnessing Adversarial Examples}},
  author={Goodfellow, Ian J and others},
  journal={arXiv preprint arXiv:1412.6572},
  year={2014}
}

@article{madry2017towards,
  title={{Towards Deep Learning Models Resistant to Adversarial Attacks}},
  author={Madry, Aleksander and Makelov, Aleksandar and Schmidt, Ludwig and Tsipras, Dimitris and Vladu, Adrian},
  journal={arXiv preprint arXiv:1706.06083},
  year={2017}
}

@incollection{kurakin2018adversarial,
  title={{Adversarial Examples in the Physical World}},
  author={Kurakin, Alexey and others},
  booktitle={Artificial intelligence safety and security},
  pages={99--112},
  year={2018},
  publisher={Chapman and Hall/CRC}
}

@inproceedings{eykholt2018robust,
  title={{Robust Physical-World Attacks on Deep Learning Visual Classification}},
  author={Eykholt, Kevin and Evtimov, Ivan and Fernandes, Earlence and Li, Bo and Rahmati, Amir and Xiao, Chaowei and Prakash, Atul and Kohno, Tadayoshi and Song, Dawn},
  booktitle={Proc. of the 31st IEEE/CVF CVPR},
  pages={1625--1634},
  year={2018}
}

@inproceedings{geiger2012we,
  title={{Are We Ready for Autonomous Driving? The KITTI Vision Benchmark Suite}},
  author={Geiger, Andreas and Lenz, Philip and Urtasun, Raquel},
  booktitle={Proc. of the 25th IEEE/CVF CVPR},
  pages={3354--3361},
  year={2012},
}

@inproceedings{xie2017adversarial,
  title={{Adversarial Examples for Semantic Segmentation and Object Detection}},
  author={Xie, Cihang and Wang, Jianyu and Zhang, Zhishuai and Zhou, Yuyin and Xie, Lingxi and Yuille, Alan},
  booktitle={Proc. of the 16th IEEE/CVF ICCV},
  pages={1369--1378},
  year={2017}
}

@inproceedings{schmalfuss2022perturbation,
  title={{A Perturbation-Constrained Adversarial Attack for Evaluating the Robustness of Optical Flow}},
  author={Schmalfuss, Jenny and Scholze, Philipp and Bruhn, Andr{\'e}s},
  booktitle={Proc. of the 17th ECCV},
  pages={183--200},
  year={2022},
  organization={Springer}
}

@inproceedings{ranjan2019attacking,
  title={{Attacking Optical Flow}},
  author={Ranjan, Anurag and Janai, Joel and Geiger, Andreas and Black, Michael J},
  booktitle={Proc. of the 17th IEEE/CVF ICCV},
  pages={2404--2413},
  year={2019}
}

@inproceedings{cheng2022physical,
  title={{Physical Attack on Monocular Depth Estimation With Optimal Adversarial Patches}},
  author={Cheng, Zhiyuan and Liang, James and Choi, Hongjun and Tao, Guanhong and Cao, Zhiwen and Liu, Dongfang and Zhang, Xiangyu},
  booktitle={Proc. of the 17th ECCV},
  pages={514--532},
  year={2022},
  organization={Springer}
}

@inproceedings{zheng2024physical,
  title={{Physical 3D Adversarial Attacks Against Monocular Depth Estimation in Autonomous Driving}},
  author={Zheng, Junhao and Lin, Chenhao and Sun, Jiahao and Zhao, Zhengyu and Li, Qian and Shen, Chao},
  booktitle={Proc. of the 37th IEEE/CVF CVPR},
  pages={24452--24461},
  year={2024}
}

@article{guesmi2023aparate,
  title={{Aparate: Adaptive Adversarial Patch for Cnn-Based Monocular Depth Estimation for Autonomous Navigation}},
  author={Guesmi, Amira and Hanif, Muhammad Abdullah and Alouani, Ihsen and Shafique, Muhammad},
  journal={arXiv preprint arXiv:2303.01351},
  year={2023}
}

@inproceedings{wong2021stereopagnosia,
  title={{Stereopagnosia: Fooling Stereo Networks With Adversarial Perturbations}},
  author={Wong, Alex and Mundhra, Mukund and Soatto, Stefano},
  booktitle={Proc. of the 35th AAAI},
  volume={35},
  number={4},
  pages={2879--2888},
  year={2021}
}

@article{cheng2021towards,
  title={{Towards Adversarially Robust and Domain Generalizable Stereo Matching by Rethinking DNN Feature Backbones}},
  author={Cheng, Kelvin and others},
  journal={arXiv preprint arXiv:2108.00335},
  year={2021}
}

@inproceedings{zagoruyko2015,
  title={{Learning to Compare Image Patches via Convolutional Neural Networks}},
  author={Zagoruyko, S. and Komodakis, N.},
  booktitle={Proc. of the 28th IEEE/CVF CVPR},
  year={2015},
  pages={4353--4361}
}

@inproceedings{zbontar2016,
  title={{Stereo Matching by Training a Convolutional Neural Network to Compare Image Patches}},
  author={Zbontar, J. and LeCun, Y.},
  booktitle={Journal of Machine Learning Research},
  volume={17},
  pages={1--32},
  year={2016}
}

@inproceedings{chang2018,
  title={{Pyramid Stereo Matching Network}},
  author={Chang, J.-R. and Chen, Y.-S.},
  booktitle={Proc. of the 31st IEEE/CVF CVPR},
  year={2018},
  pages={5410--5418}
}

@inproceedings{athalye2018synthesizing,
  title={{Synthesizing Robust Adversarial Examples}},
  author={Athalye, Anish and Engstrom, Logan and Ilyas, Andrew and Kwok, Kevin},
  booktitle={International conference on machine learning},
  pages={284--293},
  year={2018},
  organization={PMLR}
}

@inproceedings{lovisotto2021slap,
  title={{$\{$Slap$\}$: Improving Physical Adversarial Examples With $\{$Short-Lived$\}$ Adversarial Perturbations}},
  author={Lovisotto, Giulio and Turner, Henry and Sluganovic, Ivo and Strohmeier, Martin and Martinovic, Ivan},
  booktitle={Proc. of the 30th USENIX},
  pages={1865--1882},
  year={2021}
}

@inproceedings{zhu2021can,
  title={{Can We Use Arbitrary Objects to Attack Lidar Perception in Autonomous Driving?}},
  author={Zhu, Yi and Miao, Chenglin and Zheng, Tianhang and Hajiaghajani, Foad and Su, Lu and Qiao, Chunming},
  booktitle={Proc. of the 28th ACM CCS},
  pages={1945--1960},
  year={2021}
}

@inproceedings{tu2020physically,
  title={{Physically Realizable Adversarial Examples for Lidar Object Detection}},
  author={Tu, James and Ren, Mengye and Manivasagam, Sivabalan and Liang, Ming and Yang, Bin and Du, Richard and Cheng, Frank and Urtasun, Raquel},
  booktitle={Proc. of the 33rd IEEE/CVF CVPR},
  pages={13716--13725},
  year={2020}
}

@inproceedings{abdelfattah2021towards,
  title={{Towards Universal Physical Attacks on Cascaded Camera-Lidar 3D Object Detection Models}},
  author={Abdelfattah, Mazen and Yuan, Kaiwen and Wang, Z Jane and Ward, Rabab},
  booktitle={Proc. of the 28th IEEE ICIP},
  pages={3592--3596},
  year={2021},
  organization={IEEE}
}

@inproceedings{cao2021invisible,
  title={{Invisible for Both Camera and Lidar: Security of Multi-Sensor Fusion Based Perception in Autonomous Driving Under Physical-World Attacks}},
  author={Cao, Yulong and Wang, Ningfei and Xiao, Chaowei and Yang, Dawei and Fang, Jin and Yang, Ruigang and Chen, Qi Alfred and Liu, Mingyan and Li, Bo},
  booktitle={Proc. of the 42nd IEEE {S\&P}},
  pages={176--194},
  year={2021},
  organization={IEEE}
}

@inproceedings{zhu2021adversarial,
  title={{Adversarial Attacks Against Lidar Semantic Segmentation in Autonomous Driving}},
  author={Zhu, Yi and Miao, Chenglin and Hajiaghajani, Foad and Huai, Mengdi and Su, Lu and Qiao, Chunming},
  booktitle={Proc. of the 19th ACM SenSys},
  pages={329--342},
  year={2021}
}

@inproceedings{liu2024physical,
  title={{Physical Attack for Stereo Matching}},
  author={Liu, Yang and Zhai, Jucai and Ma, Chihao and Zeng, Pengcheng and Wang, Xinan and Zhao, Yong},
  booktitle={Proc. of the 1st ACM CVDL},
  pages={1--5},
  year={2024}
}

@inproceedings{thys2019fooling,
  title={{Fooling Automated Surveillance Cameras: Adversarial Patches to Attack Person Detection}},
  author={Thys, Simen and others},
  booktitle={Proc. of the 18th IEEE/CVF CVPRW},
  pages={0--0},
  year={2019}
}

@inproceedings{xu2020aanet,
  title={{Aanet: Adaptive Aggregation Network for Efficient Stereo Matching}},
  author={Xu, Haofei and Zhang, Juyong},
  booktitle={Proc. of the 33rd IEEE/CVF CVPR},
  pages={1959--1968},
  year={2020}
}

@inproceedings{wang2019pseudo,
  title={{Pseudo-Lidar From Visual Depth Estimation: Bridging the Gap in 3D Object Detection for Autonomous Driving}},
  author={Wang, Yan and Chao, Wei-Lun and Garg, Divyansh and Hariharan, Bharath and Campbell, Mark and Weinberger, Kilian Q},
  booktitle={Proc. of the 32nd IEEE/CVF CVPR},
  pages={8445--8453},
  year={2019}
}

@inproceedings{li2019stereo,
  title={{Stereo R-CNN Based 3D Object Detection for Autonomous Driving}},
  author={Li, Peiliang and Chen, Xiaozhi and Shen, Shaojie},
  booktitle={Proc. of the 32nd IEEE/CVF CVPR},
  pages={7644--7652},
  year={2019}
}

@inproceedings{dosovitskiy2017carla,
  title={{CARLA: An open urban driving simulator}},
  author={Dosovitskiy, Alexey and Ros, German and Codevilla, Felipe and Lopez, Antonio and Koltun, Vladlen},
  booktitle={Conference on robot learning},
  pages={1--16},
  year={2017},
  organization={PMLR}
}

@inproceedings{mayer2016large,
  title={{A Large Dataset to Train Convolutional Networks for Disparity, Optical Flow, and Scene Flow Estimation}},
  author={Mayer, Nikolaus and Ilg, Eddy and Hausser, Philip and Fischer, Philipp and Cremers, Daniel and Dosovitskiy, Alexey and Brox, Thomas},
  booktitle={Proc. of the 29th IEEE/CVF CVPR},
  pages={4040--4048},
  year={2016}
}

@inproceedings{zhang2019ga,
  title={{GA-Net: Guided Aggregation Net for End-To-End Stereo Matching}},
  author={Zhang, Feihu and Prisacariu, Victor and Yang, Ruigang and Torr, Philip HS},
  booktitle={Proc. of the 32nd IEEE/CVF CVPR},
  pages={185--194},
  year={2019}
}

@inproceedings{zhu2024malicious,
  title={{Malicious attacks against multi-sensor fusion in autonomous driving}},
  author={Zhu, Yi and Miao, Chenglin and Xue, Hongfei and Yu, Yunnan and Su, Lu and Qiao, Chunming},
  booktitle={Proceedings of the 30th Annual International Conference on Mobile Computing and Networking},
  pages={436--451},
  year={2024}
}

@misc{apollo,
  author       = {{Baidu}},
  title        = {Apollo: Open Autonomous Driving Platform},
  year         = {2025},
  howpublished = {\url{https://apollo.auto/}},
  note         = {Accessed: 2025-03-02}
}

@misc{waymo,
  author       = {{Waymo}},
  title        = {Waymo: Autonomous Driving Technology},
  year         = {2025},
  howpublished = {\url{https://waymo.com/}},
  note         = {Accessed: 2025-07-22}
}

@misc{mobileye,
  author       = {{Mobileye}},
  title        = {Mobileye: Advanced Driver-Assistance Systems and Autonomous Driving Solutions},
  year         = {2025},
  howpublished = {\url{https://www.mobileye.com/}},
  note         = {Accessed: 2025-05-11}
}

@inproceedings{lipson2021raft,
  title={{RAFT-Stereo: Multilevel Recurrent Field Transforms for Stereo Matching}},
  author={Lipson, Lahav and Teed, Zachary and Deng, Jia},
  booktitle={International Conference on 3D Vision (3DV)},
  year={2021}
}

@inproceedings{li2022practical,
  title={{Practical Stereo Matching via Cascaded Recurrent Network with Adaptive Correlation}},
  author={Li, Jiankun and Wang, Peisen and Xiong, Pengfei and Cai, Tao and Yan, Ziwei and Yang, Lei and Liu, Jiangyu and Fan, Haoqiang and Liu, Shuaicheng},
  booktitle={Proc. of the 35th IEEE/CVF CVPR},
  pages={9331--9340},
  year={2022}
}

@inproceedings{liang2024any,
  title={{Any-Stereo: Arbitrary Scale Disparity Estimation for Iterative Stereo Matching}},
  author={Liang, Zhaohuai and Li, Changhe},
  booktitle={Proc. of the 38th AAAI},
  volume={38},
  number={4},
  pages={3333--3341},
  year={2024}
}

\end{document}